\documentclass{article}

     \usepackage[final, nonatbib]{ml4eng_2020}

\usepackage[utf8]{inputenc} 
\usepackage[T1]{fontenc}    
\usepackage{hyperref}       
\usepackage{url}            
\usepackage{booktabs}       
\usepackage{amsfonts}       
\usepackage{nicefrac}       
\usepackage{microtype}      
\usepackage{subfigure}
\usepackage{pbox}
\usepackage{graphicx, wrapfig}
\usepackage{amssymb}
\usepackage{amsfonts}
\usepackage{amsmath}
\usepackage{amsthm}
\usepackage{amsbsy}
\usepackage{algorithm}
\usepackage{algpseudocode}

   \usepackage{xcolor}
   \definecolor{value1}{gray}{0.75}
   \definecolor{value2}{gray}{0.5}
   \definecolor{value3}{gray}{0.25}

\title{Parameterized Reinforcement Learning \\ for Optical System Optimization}

\author{%
  Heribert Wankerl\\
  University of Regensburg\\
  OSRAM Opto Semiconductors\\
  Regensburg, 93053 \\
  \texttt{heribert.wankerl@osram-os.com} \\  
  \And
  Maike L. Stern\\
  OSRAM Opto Semiconductors\\
  Regensburg, 93053 \\
   \And
  Ali Mahdavi\\
  OSRAM Opto Semiconductors\\
  Regensburg, 93053 \\
   \And
  Christoph Eichler\\
  OSRAM Opto Semiconductors\\
  Regensburg, 93053 \\
   \And
  Elmar W. Lang\\
  University of Regensburg\\
  Regensburg, 93053 \\
}

\begin{document}

\maketitle

\begin{abstract}
Designing a multi-layer optical system with designated optical characteristics is an inverse design problem in which the resulting design is determined by several discrete and continuous parameters. In particular, we consider three design parameters to describe a multi-layer stack: Each layer’s dielectric material and thickness as well as the total number of layers. Such a combination of both, discrete and continuous parameters is a challenging optimization problem that often requires a computationally expensive search for an optimal system design. Hence, most methods merely determine the optimal thicknesses of the system’s layers. To incorporate layer material and the total number of layers as well, we propose a method that considers the stacking of consecutive layers as parameterized actions in a Markov decision process. We propose an exponentially transformed reward signal that eases policy optimization and adapt a recent variant of Q-learning for inverse design optimization. We demonstrate that our method outperforms human experts and a naive reinforcement learning algorithm concerning the achieved optical characteristics. Moreover, the learned Q-values contain information about the optical properties of multi-layer optical systems, thereby allowing physical interpretation or what-if analysis.
\end{abstract}
\section{Introduction}
Modern optical systems feature complex multi-layer designs, which transmit or reflect designated parts of the wave spectrum to achieve a certain functionality \cite{Liddell1981, Macleod2010}.
Optimizing those layer stacks with respect to their optical characteristics is an inverse design problem, which covers discrete as well as continuous parameters. Namely, the total number of layers and each layer’s dielectric material properties as well as each layer’s thickness.
However, considering all these parameters results in a large number of possible designs and more particularly in a large number of designs with sub-optimal optical properties. Thus, the corresponding search space is non-convex and contains many sub-optimal local optima \cite{Liddell1981,Horst1996}. As a result, this kind of optimization problem is often solved with heuristic approaches that only optimize one parameter, as an instance the layers’ thicknesses \cite{Chang1990, Paszkowicz2013, Yang2013, Guo2014, Martin1995}, or transform the search space by considering only the discretized layer thickness values \cite{Jiang2020}. While Dobrowolski et al. \cite{Dobrowolski1996} allow to incorporate discrete and continuous parameters, their algorithm cannot incorporate dispersive materials, a prerequisite for many optical optimization problems. Other recent approaches require the pre-selection of an extensive dataset to train a differentiable surrogate model in a supervised manner \cite{Peurifoy2018}. In this work, we propose a reinforcement learning algorithm (RL, \cite{Watkins1989,vanHasselt2016}) for the optimization of multi-layer optical systems, which is based on multi-path deep Q-learning (MP-DQN, \cite{Bester2019}). Our approach allows us to incorporate all three design parameters and to operate directly in the space of so-called parameterized actions, where each discrete action is accompanied by a continuous action-parameter. Furthermore, we impose constraints on the design parameters via a Lagrangian formalism, so as to achieve a system design that features less complex structures while preserving designated reflectivity characteristics. We demonstrate our algorithm on three different optimization tasks and show that it outperforms optical system designs developed by human experts as well as a standard Q-learning algorithm \cite{Jiang2020}. In addition, many hyperparameters of MP-DQN are defined such that they have a physical correspondence regarding the proposed optical systems. Based on this, Q-value estimates are intuitively used to pursue a what-if analysis and thus investigate the behavior of a design under particular layer changes. 
\section{State of the art}
\label{state-of-the-art}
RL \cite{Sutton1998} and especially deep Q-learning have driven major advances in finding an optimal policy in many domains that allow either continuous actions \cite{Lillicrap2015} or discrete actions \cite{Mnih2013}. The combination of both, discrete and continuous actions results in parameterized action spaces \cite{Masson2016}. Recent work has found sophisticated behavior policies in domains such as 2D robot soccer \cite{Hausknecht2016, Hussein2018, Bester2019}, simulated human-robot interaction \cite{Khamassi2017} and terrain-adaptive bipedal and quadrupedal locomotion \cite{Peng2016}. In general, the approaches to solving tasks that include parameterized actions are two-fold. First, hierarchical techniques separate the optimization of discrete actions and continuous action-parameters by iteratively alternating between them during optimization \cite{Masson2016, Khamassi2017}. Therefore, they omit an exchange of information between the policies for discrete and continuous actions, respectively. Second, some recent work focuses on transforming the parameterized actions into continuous \cite{Hausknecht2016} or discrete ones \cite{Jiang2020}. Here, the interaction between continuous and discrete actions is not exploited. Hence, by construction, these concepts are not suitable to represent the intrinsic information contained in parameterized action spaces. However, Xiong et al. \cite{Xiong2018} adapted deep Q-learning (DQN, \cite{vanHasselt2016}) to parameterize each discrete action with a continuous value, thereby incorporating interactions between them. The proposed path-DQN (P-DQN) allows policy optimization directly in a parameterized action space. Bester et al. \cite{Bester2019} suggested so-called multi-path DQN (MP-DQN) based on their assumption that P-DQN implements the Bellman equation for parameterized action spaces incongruously. Based on MP-DQN, we propose an algorithm for solving inverse design problems that include parameterized actions. Namely, we optimize optical systems while avoiding unphysical assumptions and sticking closely to the physical domain. For instance, each discrete material choice is parameterized by a continuous thickness value. A sequence of such design choices results in a multi-layer optical system. To achieve discriminability of different optical systems in terms of reward, we introduce a domain-agnostic exponential transformation that can be adapted to other optimization tasks, e.g. when a reconstruction error should be minimized.
\begin{figure}[b]
\vspace{-23pt}
\centering
\includegraphics[width=8cm]{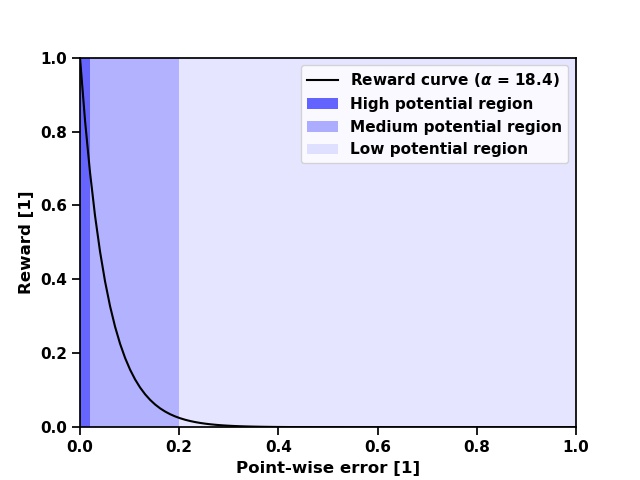}%
\caption{Illustration of the mapping between error and reward, highlighting the regions that divide the search space.}
\label{fig:reward_curve}
\end{figure}

\section{Optical systems and the inverse design problem}
\label{optics}
In this work, the design of an optical system is specified by three parameters, starting with the total number $L \in \mathbb{N}$ of layers in the layer stack. Each of these consecutive layers consists of a material with a certain refractive index and a specified thickness. Thus, we can encode all parameters of a layer as a vector $\mathbf{n} \in \mathbb{C}^L$ of refractive indexes and a vector $\mathbf{t} \in \mathbb{R}^L$ of thickness values, respectively. Based on a simulation, the observed reflectivity $R_{\lambda, \varphi}\left( \mathbf{n}, \mathbf{t} \right)$ is obtained as a function of the design parameters $\mathbf{n}$ and $\mathbf{t}$ as well as the wavelength $\lambda$ and the incident angle $\varphi$ of the incoming light. Here, a light-emitting diode functions as a light source that emits an unpolarized electromagnetic spectrum at different angles. We thus get a vector of reflectivity values 
$ \mathbf{R}\left( \mathbf{n}, \mathbf{t} \right) = \left( R_{\lambda , \varphi}\left( \mathbf{n}, \mathbf{t} \right) \vert \lambda \in \Lambda, \varphi \in \Phi \right)$, 
where $\Lambda, \Phi \subset \mathbb{R}$ denote discrete and compact sets of wavelengths and incidence angles of the emitted radiation, respectively. Based on the intended application of an optical system, the design is required to feature a target reflectivity vector
$ \mathbf{T} = \left( T_{\lambda, \varphi} \vert \lambda \in \Lambda, \varphi \in \Phi \right)$.
Therefore, we can propose an objective function
\begin{equation}
\label{error}
F\left(\mathbf{n}, \mathbf{t}, \mathbf{T} \right) =
- \dfrac{1}{\vert \Phi \vert \cdot \vert \Lambda \vert} \sum\nolimits_{\varphi \in \Phi} \sum\nolimits_{\lambda \in \Lambda}
\vert R_{\lambda, \varphi}\left( \mathbf{n}, \mathbf{t} \right)
- T_{\lambda, \varphi} \vert^2 - \dfrac{\mu}{L} \cdot \sum\nolimits_{l=1}^L t_l \text{   ,     } \mu > 0 
\end{equation}
that we aim to maximize. Here, the first summand computes the mean squared error (MSE) between a given and a target reflectivity curve. The multiplier $\mu$ in the second addend, a Lagrangian term, introduces regularization, which punishes complex design suggestions. Complexity here refers to the number of layers and layer thicknesses.
However, using this constrained object function as a reward signal for the RL algorithm results in barely differentiable rewards for designs with reflectivity values close to the target reflectivity. This effect may be attributed to the quadratic form of equation \eqref{error}, which yields high, but nearly constant values for near-optimal designs. We address this shortcoming by introducing an exponential transformation $ r \equiv \exp\left(\alpha \cdot F \right) \text{, } \alpha>0$, which scales the observed reward $r$ between $0.01$ and $1$. Here, $\alpha$ is an empirically determined scaling hyperparameter, as explained in appendix~\ref{appendix:B}. As illustrated in figure~\ref{fig:reward_curve}, the reward function now emphasizes the differences in near-optimal system designs while design options with undesirable optical responses are still assigned a low reward. Following the Bellman equation, the discriminability of the rewards is directly imparted to the estimated Q-values, which in turn evaluate the given states. As a result, decision making and learning are improved in general.
\begin{figure}[b]
\vspace{-0.3cm}
      \begin{algorithm}[H]
        \caption{MP-DQN for inverse design optimization} \label{alg:MP-DQN}
        \begin{algorithmic}[1]
          \State Initialize $\theta, \theta', E, L, \mathcal{D}, \tau$
          \For{$e = 1:E$}
          \State Initialize $s_0$ (with zeros) and adapt $\epsilon$
          \For{$i = 0:(L-1)$}
          \State With probablity $\epsilon$ select random action $(n_i, t_i)$
          \State Otherwise select $\smash{a_i = (n_i, t_i) = \text{argmax}_{a'} ( \widehat{Q}( s_i, a' \vert \theta ))}$
          \State Stack layer  $(n_i, t_i)$ and observe $r_i, s_{i+1}$
          \State Store transition $(s_i, a_i, r_i, s_{i+1})$ in $\mathcal{D}$
          \EndFor
          \State Sample random mini-batch $\mathcal{B} \subset \mathcal{D}$ of transitions $\lbrace \left( s_j, a_j, r_j, s_{j+1} \right) \rbrace_{j}$
          \State For each transition compute $\smash{y = r_j + \gamma \cdot \text{max}_{a'} (\widehat{Q} \left( s_{j+1}, a' \vert \theta' \right) )}$
          \State Compute loss $\mathcal{L} = \sum_{\mathcal{B}}( y - \widehat{Q} \left( s_j, a_j \vert \theta \right) )^2$ 
          \State Perform gradient descent on $\theta$ following Bester et al. \cite{Bester2019}
          \If{target network update}
          	\State Update $\theta' $ using Polyak averaging $\theta' \gets \tau \cdot \theta + (1-\tau) \cdot \theta'$
          \EndIf
          \EndFor
        \end{algorithmic}
      \end{algorithm}
\end{figure}
\newpage
\section{Reinforcement learning for optimization in parameterized action spaces}
\label{rl}
In RL, an agent aims to maximize a reward signal that is calculated with respect to the environment's current state. Such a state can be described as a concatenated set $s_i = \lbrace \mathbf{n}, \mathbf{t} \rbrace \subset \mathcal{S}, $ where $ i $ is the current episode's step number and $\mathcal{S}$ denotes the set of possible states. At the beginning of each of the $E \in \mathbb{N}$ episodes, all entries of the vectors  $\mathbf{n}$ and $\mathbf{t}$ are set to zero. As stated in algorithm~\ref{alg:MP-DQN}, the agent successively executes parameterized actions $a_i = \left( n_i, t_i \right) \in N \times T$, which determine the refractive index $ \mathbf{n}_i $ and the thickness $ \mathbf{t}_i $ of the current layer $i \leq L$.
Instead of choosing $ \mathbf{n}_i $ and $ \mathbf{t}_i $, the agent can also terminate the episode and hereby determine the total number of layers $ l $ of the current optical system, such that $l \leq L$. The parameterized action space becomes $ \mathcal{A} = \lbrace a = \left( n, t \right) \vert n \in N , t \in T \rbrace $. Obviously, the pre-definition of the sets of possible thickness values $T \subset \mathbb{R}^+$ and available refractive indexes $N \subset \mathbb{C}$ allows to impose additional hard constraints on the optimization. After an episode is terminated, either by the agent's choice or by reaching the maximum number of layers $ L $, the optical system's reflectivity curve is simulated. Based on this reflectivity a reward is assigned, as explained in section \ref{optics}. In order to minimize costly calls to the simulation software, each of the non-terminal states is assigned a zero reward. Because these so-called delayed rewards impede Q-value approximation, we rate non-terminal states recursively using an $ l $-step return,
$
r_{i-1} \leftarrow \gamma \cdot r_{i}, \text{ , } 0<i \leq l
$,
where $r_{l} \equiv r$ is the final reward and $\gamma =0.95$ is the discount factor for the future reward.\par 
The described formalism allows us to interpret the problem as a parameterized action Markov decision process  $\left( \mathcal{S}, \mathcal{A}, \mathbb{P}, r, \gamma \right)$ (PAMDP, \cite{Masson2016}), where $\mathbb{P}\left( s_{i+1}\vert s_i\right)$ is the Markov state transition probability function. Each transition in this process gets stored in a replay memory $\mathcal{D}$, as a tuple of the current state~$s_i$, the taken action $a_i$, the subsequent state $s_{i+1}$, and the $l$-step return $r_i$. Using MP-DQN, the collected data and the Bellman equation are used to approximate the Q-values
\begin{equation}
\label{eq:expectation}
Q(s_i, a_i) = \mathbb{E}_{r_{i}, s_{i+1}} [
r_{i} + \gamma \cdot \max_{a_{i+1}} Q\left( s_{i+1}, a_{i+1} \right) \vert s_{i}, a_{i} ]
\end{equation}
that are the expected future rewards given a current state and a particular parameterized action. 
As a result, the optimal policy $ \pi: s \mapsto \text{argmax}_{a'} \smash{\widehat{Q}(s, a')} $ is given by taking actions $a$ corresponding to maximum Q-value estimates $\smash{\widehat{Q}(s, a) \approx Q(s, a)}$ in a particular state $s$. To approximate the Q-values, we implement a sequence of deep neural networks $f$ and $g$ with joint parameterization~$\theta$. Briefly explained, we estimate possible thickness values for each material available given the current state by the network $ g: \mathcal{S} \mapsto T^{\vert N \vert} $ that features $\vert s \vert$ input nodes and $\vert N \vert$ output nodes. Each output node corresponds to a material in $N$ and suggests the thickness value of the next layer to stack if the respective material is chosen. Which material is actually chosen is based on the multi-path policy evaluation $f(s, g(s) \vert \theta)$, with $\vert N \vert +1$ outputs. Each output value represents a Q-value estimate, $\smash{\widehat{Q}(s, a \vert \theta) \equiv \widehat{Q}(s, a)}$, for the associated parameterized action while taking into account both, the current state $s$ and the suggestions for thickness values $g(s)$. Note that there is one additional node, which represents the action that terminates an episode. We can summarize that MP-DQN extends the DQN algorithm so as to solve PAMDPs by considering network $g$ as an intermediate continuous actor and network $f$ as an approximator of Q-values, thus functioning as a discrete actor.\par
As in common DQNs, the successively collected data is highly correlated and its distribution varies due to policy adaption during optimization. This violates the assumption of independent and identically distributed data for neural network training. Hence, to stabilize policy optimization we introduce a target network \cite{vanHasselt2016} and a replay memory $\mathcal{D}$ \cite{Mnih2013}, where sampling from $\mathcal{D}$ breaks the correlation between data generated by the same trajectory. The target and policy network feature two hidden layers with $256$ nodes each. As outlined in algorithm~\ref{alg:MP-DQN}, after each episode and entailed $l$-step return calculation, the policy network parameterization $\theta$ is updated with a learning rate of $0.001$. The target network parameterization $\theta'$ is updated every ten episodes using Polyak averaging, with $\tau=0.01$. The replay memory was adapted for optical design optimization by implementing a non-uniform random drawing of training batches, so-called prioritization \cite{Schaul2016}. The probability of choosing a particular transition from the replay memory is determined by applying the softmax function to the losses of transitions. Thus, transitions that correspond to misestimated Q-values have a higher probability of getting sampled. Another important aspect of optimization algorithms in general is the exploration-exploitation trade-off that is implemented through an $\epsilon$-greedy policy in this work. We adapt $\epsilon \in [\epsilon_{final}, 1]$ before each episode. Beginning from $\epsilon=1$, we exponentially reduce $ \epsilon $ by a factor of $0.997$ until $\epsilon = \epsilon_{final}$, such that $\smash{( 1 - \epsilon_{final} )^L \approx 0.3}$ holds. This turned out to be an adequate long-term trade-off between exploration and exploitation as the agent can design an optical system in $3$ out of $10$ episodes without any random exploration. Note that RL is employed to solve an optimization problem. Thus, convergence of the policy is not intended, because this would result in proposing the same optical system again and again without any additional information gain.
\section{Experiments}
\begin{figure}[t]
\vspace{-1.0cm}
\subfigure[Task $1$]{\includegraphics[width=0.49\textwidth]{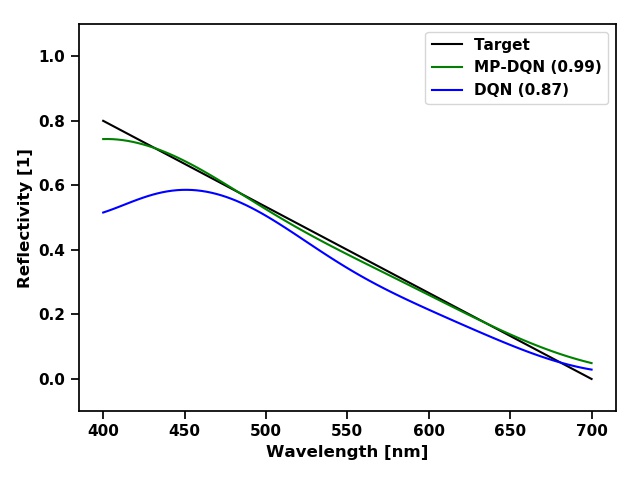}}
\subfigure[Task $2$]{\includegraphics[width=0.49\textwidth]{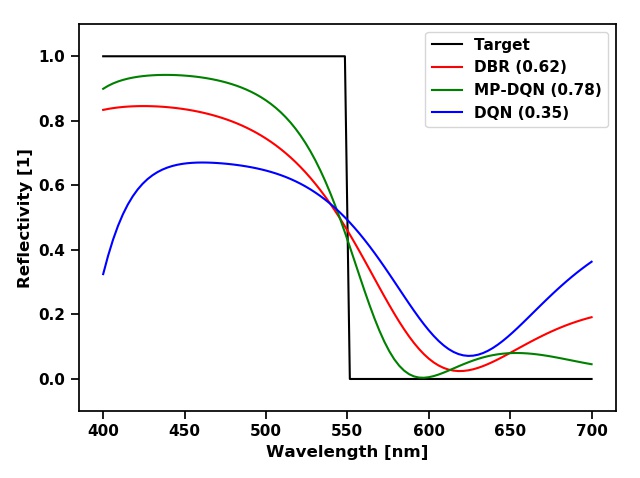}}
\caption{Illustration of the target and reflectivity curves that correspond to the highest obtained reward using MP-DQN (ours) and DQN. The achieved reward is denoted in brackets. In addition, the reflectivity curve obtained by a distributed Bragg reflector (DBR, see appendix \ref{appendix:A}) is visualized for task $2$. We set $\alpha = 18.42$ and $\mu = 0$ in order to compute the reward based on equation \eqref{error}.}
\label{fig:task12}
\vspace{-0.2cm}
\end{figure}

To analyze our MP-DQN approach, we perform optimization on three different tasks, as stated in table \ref{tab:experiments}. To realize the extend of the corresponding search spaces, we can approximate the total number of possible states to be $\smash{\vert \mathcal{S} \vert = \sum_{l=1}^{L} \vert T \vert ^l \cdot \vert N \vert \cdot ( \vert N \vert -1 )^{l-1}}$, if we assume discrete layer thicknesses from $0 - 150 \, nm$ in steps of $0.1 \, nm$ resulting in a total number of $\vert T \vert = 1500$ thickness values \cite{Jiang2020}. We compare our experimental results to optical systems designed by human experts and another Q-learning algorithm \cite{Jiang2020}, henceforth referred to as DQN algorithm. However, to enhance comparability between our approach and the DQN algorithm, we enabled the latter to not only optimize over layer thicknesses but also for layer material. Nevertheless, contrary to our approach, DQN operates on discretized thickness values and a pre-defined stack consisting of a fixed number of layers. Therefore, DQN's design initialization was set to a random but fixed layer stack at the beginning of each of the $200$ episodes, which cover $250$ steps each. We run DQN ten times and report the reflectivity curves corresponding to the highest achieved rewards for tasks $1$ and $2$. After running our approach once for $10,000$ episodes with $L$ steps each, we compare the results of our approach and the DQN algorithm. Figure~\ref{fig:task12} reveals that we distinctively outperform DQN not only in terms of achieved best rewards that were improved by at least $20\%$ for task $1$ and $2$: Whereas our approach employs $10,000$ simulation calls, DQN relies on one simulation call per step resulting in $50,000$ simulation calls in each run. Moreover, the same figure states that MP-DQN achieves an even higher reward compared to a distributed Bragg reflector (DBR, see appendix \ref{appendix:A}), which is a physically deduced solution for task $2$.\par
\begin{figure}[t]
\vspace{-1cm}
\centering
	\includegraphics[width=.99\linewidth]{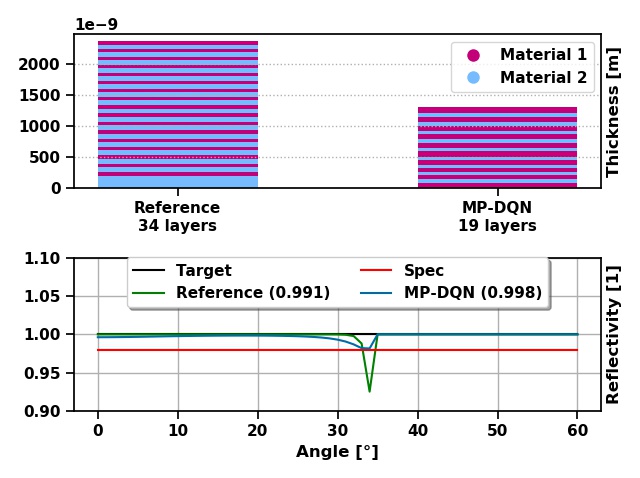}%
    \caption{Task 3. On top, the reference design and the design obtained by MP-DQN is depicted. The bottom illustration depicts the target and specification reflectivity as well as the averaged reflectivities for considered wavelengths over angle.}
    \label{fig:task3}
\end{figure}

\begin{table*}[b]
\centering
\footnotesize
\makebox[\linewidth]{
\begin{tabular}{c|lcccccc}
	ID  & $\bold{T}$ & $\Lambda [nm]$ & $\Phi [^\circ]$ & $L$ & $\vert \mathcal{S} \vert$  & $\vert N \vert$ \\ \hline	
	1  &    $T_{\lambda, \varphi} = \nicefrac{1}{375} \cdot \lambda - \nicefrac{16}{15} $ & $[400, 700]$ & $\lbrace 0 \rbrace$ & $8$ & $2.24 \cdot 10^{29}$ & 4\\
	2 &     $T_{\lambda, \varphi} = \nicefrac{1}{2} \cdot \left[ 1- \text{tanh} \left( \lambda - 550 \right) \right]$ & $[400, 700]$ & $\lbrace 0 \rbrace$ & $8$ & $2.24 \cdot 10^{29}$ & 4\\
	3  &    $T_{\lambda, \varphi} = 1.0 $ & $[445, 455]$ & $[0, 60]$ & $34$ & $ 1.94 \cdot 10^{108}$ & 2\\
\end{tabular}
}
\caption{Summary of the tasks including their target curves $\mathbf{T}$, considered wavelengths $\Lambda$ and incident angles $\Phi$. $L$ denotes the maximum number of layers placed, $\vert N \vert$, and $\vert \mathcal{S} \vert$ are the number of available materials and the approximate number of states of the resulting PAMDP, respectively.}
\label{tab:experiments}
\end{table*}
\textbf{Constrained Optimization} \newline
To control the complexity of the designs created by our MP-DQN approach, we run task 1 again, using a constrained optimization by setting $\mu = 0.1$ in equation~\eqref{error}. When comparing designs that achieve the same unconstrained reward of approximately $0.99$, performing constrained optimization yields a distinctively thinner design with a total thickness of $503.7\,nm$, whereas the unconstrained approach ($\mu = 0.0$) suggests $598.2\,nm$. Note that as the constrained reward features an additional non-zero term, the comparison of unconstrained and constrained reward is invalid. Thus, we report and compare unconstrained rewards for both cases. Due to the convincing results, we apply the same Lagrangian multiplier $\mu = 0.1$ to optimize task $3$. We compare the constrained optimization result with a reference design that consists of $34$ layers and was developed by human experts. As shown in figure~\ref{fig:task3}, we outperform the reference and satisfy the specification (Spec, red line), using only $19$ layers, with $1307.1\,nm$ thickness in total. Practically, this reduction in complexity not only decreases production costs but also reduces optical absorption losses in the stack.\par
\textbf{Review from a physical point of view} \newline
A physicist's intuition about solving task $2$ corresponds to a DBR. Here, our approach coincides with the respective material configuration---except for the last layer. As table \ref{tab:qvalues} shows, the agent places material $3$ instead of further alternating between materials $ 1 $ and $4$. Inspired by the finding that material $4$ surprisingly features the lowest Q-value, we analyzed Q-values in terms of optical characteristics. Therefore, we compare the Q-value estimation $\smash{\widehat{Q}(s_i,a_i)}$ of each transition~$i$ of an episode with respect to the optical characteristics of the underlying parameterized action $a_i = (n_i, t_i)$ given the same state $s_i$. The first optical characteristic that we consider is the refractive index $n_i$, the second characteristic is the resulting optical path length $p_i = n_i \cdot t_i$. Interestingly, table~\ref{tab:qvalues} indicates that the functional dependencies $\smash{\widehat{Q}(s_i, n_i) \approx \widehat{Q}(s_i, a_i)}$ shows monotonic and in general convex behavior and non-convex behavior in case of $\smash{\widehat{Q}(s_i, p_i) \approx \widehat{Q}(s_i, a_i)}$ for a fixed state $s_i$, respectively. These relations suggest that the relative order of the Q-value estimates is mainly based on the refractive indexes rather than thicknesses that are associated with an action. Moreover,  as convexity prohibits the existence of local optima aside from the global optimum, Q-values seem to validly reflect relative adequacy of actions in terms of their associated refractive indexes in a particular state.\par
In addition, the expected future reward provides further physical understanding by conducting a what-if analysis. Namely, Q-values are interpreted as estimations of $l$-step returns and thus design behavior, e.g. when a particular layer is changed. This was validated by following the optimal policy until layer $i$, taking a non-optimal parameterized action, and then following the optimal policy again until the terminal state. After conducting this for every possible parameterized action, the observed $l$-step returns $r_i$ were collected in table \ref{tab:qvalues}. These results indicate that the influence of a design choice on the obtained $l$-step return is identified by the Q-values. Thus, engineers can infer physical knowledge, e.g. investigating where and why the optimal optical system deviates from a physical intuition as exemplified above for task $2$. We elucidate the acquired insights about convexity and the what-if analysis in appendix \ref{appendix:C} while also providing information about the learning dynamics.\par
\begin{table*}[t]
\vspace{-0.5cm}
\centering
\footnotesize
\makebox[\linewidth]{
\begin{tabular}{cc|c|cccccccc|}
	Mat. & $\text{Re}(\mathbf{n}_i)$ & Layer $i$  & 1 & 2 & 3 & 4 &5 &6 & 7 &8 \\  \hline	\hline
	1 & $1.457$ & \pbox{20cm}{\vspace{10pt}  $\smash{\widehat{Q}}$ \\ $p_i$ \\ $r_i$ \vspace{2pt}}
	& \pbox{20cm}{ \textbf{0.501} \\ 0.580 \\ 0.544  }  
	& \pbox{20cm}{ \textbf{\textcolor{value2}{0.297}} \\ 0.380 \\ 0.256}
	& \pbox{20cm}{ \textbf{0.551} \\ 0.735 \\ 0.603}
	& \pbox{20cm}{ \textbf{\textcolor{value1}{0.423}} \\ 0.312 \\ 0.429}
	& \pbox{20cm}{ \textbf{0.631} \\ 0.790 \\ 0.668}
	& \pbox{20cm}{ \textbf{\textcolor{value1}{0.514}} \\ 0.613 \\ 0.493}
	& \pbox{20cm}{ \textbf{0.647} \\ 1.199 \\ 0.741}
	& \pbox{20cm}{ \textbf{\textcolor{value2}{0.509}} \\ 0.376 \\ 0.499} \\ \hline
	2 & $1.645$ & \pbox{20cm}{\vspace{10pt}  $\smash{\widehat{Q}}$ \\ $p_i$ \\ $r_i$ \vspace{2pt}}  
	& \pbox{20cm}{ \textbf{\textcolor{value3}{0.388}} \\ 0.636 \\ 0.414}
	& \pbox{20cm}{ \textbf{\textcolor{value1}{0.270}} \\ 0.834 \\ 0.257}
	& \pbox{20cm}{ \textbf{\textcolor{value3}{0.506}} \\ 0.742 \\ 0.485}
	& \pbox{20cm}{ \textbf{\textcolor{value2}{0.484}} \\ 0.575 \\ 0.477}
	& \pbox{20cm}{ \textbf{\textcolor{value3}{0.596}} \\ 0.939 \\ 0.619}
	& \pbox{20cm}{ \textbf{\textcolor{value2}{0.527}} \\ 0.551 \\ 0.605}
	& \pbox{20cm}{ \textbf{\textcolor{value2}{0.517}} \\ 0.279 \\ 0.568}
	& \pbox{20cm}{ \textbf{\textcolor{value3}{0.536}} \\ 0.357 \\ 0.513} \\ \hline
	3 & $1.860$ & \pbox{20cm}{\vspace{10pt}  $\smash{\widehat{Q}}$ \\ $p_i$ \\ $r_i$ \vspace{2pt}}
	& \pbox{20cm}{ \textbf{\textcolor{value2}{0.316}} \\ 0.663 \\ 0.303}  
	& \pbox{20cm}{ \textbf{\textcolor{value3}{0.362}} \\ 0.703 \\ 0.337}
	& \pbox{20cm}{ \textbf{\textcolor{value2}{0.416}} \\ 0.967 \\ 0.427}
	& \pbox{20cm}{ \textbf{\textcolor{value3}{0.544}} \\ 0.661 \\ 0.567}
	& \pbox{20cm}{ \textbf{\textcolor{value2}{0.586}} \\ 1.273 \\ 0.566}
	& \pbox{20cm}{ \textbf{\textcolor{value3}{0.578}} \\ 0.609 \\ 0.559}
	& \pbox{20cm}{ \textbf{\textcolor{value3}{0.612}} \\ 1.473 \\ 0.589}
	& \pbox{20cm}{ \textbf{0.714} \\ 0.313 \\ 0.780} \\ \hline
	4 & $2.327$ & \pbox{20cm}{\vspace{10pt}  $\smash{\widehat{Q}}$ \\ $p_i$ \\ $r_i$ \vspace{2pt}}
	& \pbox{20cm}{ \textbf{\textcolor{value1}{0.232}} \\ 0.793 \\ 0.182} 
	& \pbox{20cm}{ \textbf{0.539} \\ 0.694 \\ 0.573}
	& \pbox{20cm}{ \textbf{\textcolor{value1}{0.339}} \\ 1.669 \\ 0.294}
	& \pbox{20cm}{ \textbf{0.651} \\ 0.792 \\ 0.634}
	& \pbox{20cm}{ \textbf{\textcolor{value1}{0.457}} \\ 1.647 \\ 0.493}
	& \pbox{20cm}{ \textbf{0.682} \\ 0.665 \\ 0.703}
	& \pbox{20cm}{ \textbf{\textcolor{value1}{0.559}} \\ 1.578 \\ 0.575}
	& \pbox{20cm}{ \textbf{\textcolor{value1}{0.395}} \\ 2.296 \\ 0.433} \\ \hline
\end{tabular}
}
\caption{Each row represents an available material (Mat.), where $\text{Re}(\mathbf{n}_i)$ denotes the real parts of the associated refractive indexes. Each column $1-8$ corresponds to a layer $i$. The first sub-row in each column contains the estimated Q-values $\smash{\widehat{Q}}$ while following the optimal policy for task $2$. The grayscale values indicate relative differences in the magnitude of Q-values in each column. The second sub-row in each column contains the optical path length $p_i$, the third sub-row the $l$-step return $r_i$ resulting if a particular action was taken and we follow the optimal policy in each (other) state.}
\label{tab:qvalues}
\vspace{-0.5cm}
\end{table*}
\textbf{Conclusion} \newline
In this work, we introduce a novel method to optimize optical system designs that require discrete as well as continuous parameters, using multi-path deep Q-learning (MP-DQN). Our contributions are three-fold: First, we used MP-DQN to address constrained inverse design problems, by formulating them as parameterized action Markov decision processes. Notably, our approach abandons the unphysical discretization of continuous variables and as a result, distinctively outperforms other methods. Second, we developed a constrained objective function to compute rewards based on an exponential transformation. The resulting reward signal becomes differentiable, which eases the agent’s policy optimization and decision making. Moreover, it enables us to control the complexity of the system designs, which reduces production costs and decreases optical absorption losses. Finally, we performed a what-if analysis based on Q-value estimates and thereby demonstrate how optical engineers can gain physical insights from the estimated Q-values.
\begin{ack}
This research was conducted at OSRAM Opto Semiconductors GmbH.\newline
We thank our colleagues for their assistance regarding optical expertise and editorial input, especially Daniel Grünbaum for various comments that greatly improved this manuscript.\par The authors declare no competing interests.
\end{ack}
\bibliographystyle{abbrv} 
\bibliography{PF-MDS-EMT_v9}
\appendix
\section{Filter construction using distributed Bragg reflectors}
\label{appendix:A}
A distributed Bragg reflector (DBR) is an efficient optical reflector that consists of alternating thin films of materials with different refractive indexes. Basically, a DBR is determined by two thickness values $t_1, t_2 \in \mathbb{R}^+$ and real refractive indexes $n_1,n_2 \in \mathbb{R}^+$, where $n_1 < n_2$ holds. Task $2$ of table \ref{tab:experiments} corresponds to a high-pass filter in the wavelength domain, because wavelengths lower than $550\, nm$ should be reflected. To obtain a physically deduced filter and thus a solution to task $2$, we can use a DBR \cite{Macleod2010}. Here, the wavelength width $\Delta \lambda$ of the stopping band can be computed with respect to the center wavelength $\lambda_0$ of the stopping band. In addition, we want the stopping band to end at $550\, nm$ and set $n_1 = 1.457$ and $n_2=2.327$. The obtained linear equation system
\begin{align*}
\Delta \lambda &= \frac{4 }{\pi} \cdot \lambda_0 \cdot \text{arcsin} \left\lvert \frac{n_2 -n_1}{n_2 + n_1} \right\rvert \\
\lambda_0 + \Delta \lambda &= 550 \, nm
\end{align*}
can be solved yielding $\lambda_0 = 424.59 \, nm$. The resonance condition for first order constructive interference
$n_1 \cdot t_1 = n_2 \cdot t_2 = \nicefrac{\lambda_0}{4}$ yields $t_1 = 72.85 \, nm$ and $t_2 = 45.62 \, nm$. To obtain an $8$-layer DBR of $473.88 \, nm$ total thickness, we repeatedly stack these two layers four times.

\section{Impact of reward transformation and Q-value reliability}

\begin{figure}[b]
\centering
\begin{minipage}[t]{0.49\linewidth}
\centering
	\includegraphics[width=.99\linewidth]{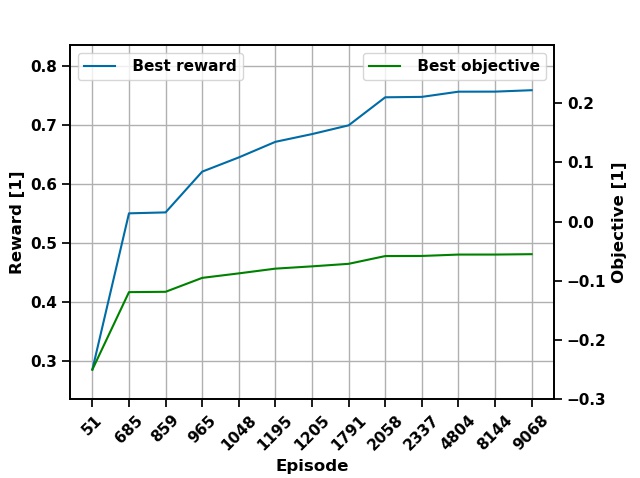}%
    \caption{The best obtained objective \eqref{error} and reward over episode of its achievement for $\alpha=18.42$ reveals the higher discriminability of designs during training. Axis limits are chosen such that the absolute length of the axes of reward and error coincide.}
    \label{fig:reward_in_practice}
\end{minipage}
\hfill
\begin{minipage}[t]{0.49\linewidth}
\centering
\includegraphics[width=.99\linewidth]{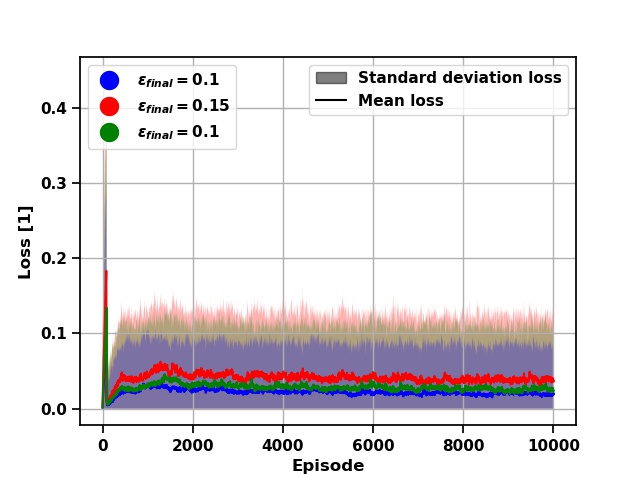}%
	\caption{Illustration of the standard deviation of and mean value of the computed loss over episode. We investigated different configurations of $\epsilon_{final}$. Note that we omitted the loss-weighted sampling of mini-batches in one case (green).}
\label{fig:dynamics_loss}
\end{minipage}
\end{figure}
\label{appendix:B}
Following the optimal policy, which leads to an optimal optical system, relies on an accurate Q-value estimation for as many state-action pairs as possible. Moreover, to ease decision making, the Q-value estimates for particular parameterized actions should be as distinguishable as possible. This condition does not apply if the rewards of more and more improved designs remain almost constant, because equation \eqref{eq:expectation} and its implementation in algorithm \ref{alg:MP-DQN} reveal that in such a case the Q-value estimates will be almost constant, too. On the other hand, many regions in the design search space are completely inadequate for solving a given task and should be assigned with very small reward. This is why we introduce a dedicated reward transformation, which relies on a hyperparameter $\alpha>0$ and is illustrated in figure \ref{fig:reward_curve}. The hyperparameter is computed by 
$$
\alpha = - \frac{1}{\eta} \cdot \text{ln} \left( \frac{\beta_1}{\beta_2} \right) = 18.42,
$$
where $\beta_1 = 0.01$ and $\beta_2=1.0$ are the lower and upper bound hyperparameters for the reward, respectively. The empirical mean value $\eta=0.25$ of equation \eqref{error} is computed based on $1,000$ randomly drawn optical systems. The impact of this transformation regarding task $2$ is illustrated in figure \ref{fig:reward_in_practice}.\par
\begin{figure}[t]
\centering
\begin{minipage}[t]{0.49\linewidth}
\centering
	\includegraphics[width=.99\linewidth]{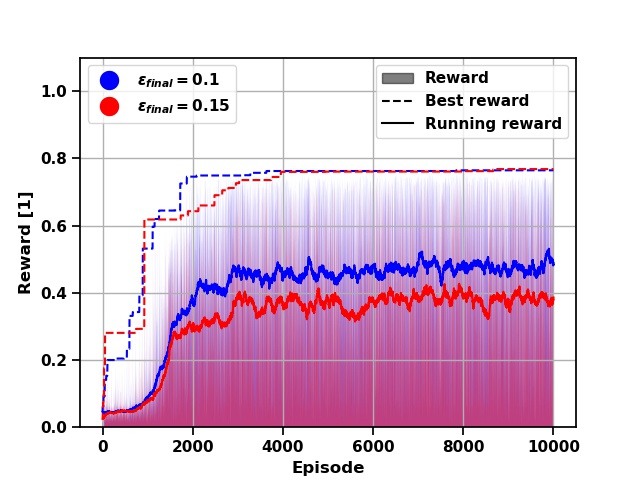}%
	\caption{Illustration of obtained reward (filled area) and running reward (solid line) over episode. The best obtained reward is indicated by dashed lines for two configurations of $\epsilon_{final}$.}
\label{fig:dynamics_reward}
\end{minipage}
\hfill
\begin{minipage}[t]{0.49\linewidth}
\centering
\includegraphics[width=.99\linewidth]{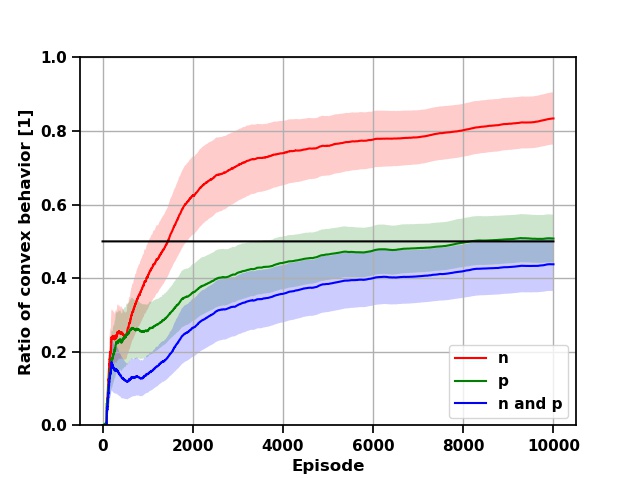}%
	\caption{Step ratio of convex behavior of Q-value approximation in terms of refractive index ($n$), optical path length ($p$), and corresponding coincidence ($n$ and $p$) over episode, respectively.}
\label{fig:dynamics_convex}
\end{minipage}
\end{figure}
It is often not discussed that the approximation of Q-values can be monitored during policy optimization. In figure \ref{fig:dynamics_loss}, we depict the mean value and standard deviation of the loss $\mathcal{L}$ for task $2$, which is computed every episode according to algorithm \ref{alg:MP-DQN}, based on the entire data in the replay memory~$\mathcal{D}$. Unsurprisingly, in the beginning, the loss is high, because the training, which is based on batches of size $128$, starts when the replay memory of total size $5,000$ contains an initial number of $500$ transitions. This prevents the neural network parameterizations from being biased due to very limited data in the early training phase. Moreover, the impact of different final exploration probabilities $\epsilon_{final}$ and the effect of prioritization is observable. Whereas a higher value for $\epsilon_{final}$ implies more exploration of unknown regions of the search space and thus uncertainty in the underlying Q-value estimation, prioritization reduces the standard deviation of loss values by preferring misestimated transitions for sampling into the mini-batches used for training. Monitoring the approximation of Q-values in the replay memory can function as an indicator in many respects: Whether to initiate more exploration in case of overfitting or whether the engineers can trust a Q-value approximation in general or should adapt their hyperparameters.
\section{Learning dynamics}
\label{appendix:C}
In addition to the loss, we also tracked $l$-step returns and eventually achieved rewards for each episode. Figure \ref{fig:dynamics_reward} depicts these measures for two different values of the final exploration probability $\epsilon_{final}$ solving task $2$. As expected, we achieve higher running rewards with lower $\epsilon_{final}$. In addition and more importantly, the best obtained reward remains nearly stable in both cases although the best proposed optical design differs due to various local optima in the search space. Finally, we investigated how the functional behavior of Q-values evolves during optimization. As a Q-value is related to a parameterized action in step $i$, we characterize the latter by either the refractive index $n_i$ or the optical path length $p_i$. We track in each step $i \leq L$ of an episode whether the estimated Q-values are convex in terms of refractive index $\smash{\widehat{Q}(s_i, n_i) \approx  \widehat{Q}(s_i,a_i)}$ or optical path length $\smash{\widehat{Q}(s_i, p_i) \approx  \widehat{Q}(s_i,a_i)}$ given the same state $s_i$. Based on the tracked data, the ratio between convex estimates and the total number of steps in each episode is calculated. Figure \ref{fig:dynamics_convex} illustrates how the running mean and standard deviation of these ratios evolve over episodes. Here, an additional measure is covered: The ratio of steps in each episode that were convex with respect to both, refractive index and optical path length. As we estimate four material-related Q-values per step, the combinatorially deduced probability for the estimates to show convex behavior is $50 \%$. This regime of randomness is indicated by the black rule in figure \ref{fig:dynamics_convex}. The running mean and standard deviation of ratios were computed based on Welford's online algorithm \cite{Welford1962}. Although the optical path length intuitively gives a more encompassing optical information about a parameterized action, the ratios of convex behavior based on refractive indexes (red rule) of $0.6 - 0.8$ are higher than for optical path lengths that are around random guessing at $0.5 -0.6$. Moreover, a comparison indicates that if the Q-value estimates are convex in terms of optical path length (green rule), they are also convex in terms of refractive indexes and thus both optical characteristics (blue rule). In general, coincident convexity in terms of both optical characteristics cannot be proven. But it seems that the Q-value estimates reflect some optical characteristics and thus contain information about the optical similarity of corresponding parameterized actions.

\end{document}